\documentclass[10pt,twocolumn,letterpaper]{article}

\usepackage[pagenumbers]{cvpr} 

\definecolor{cvprblue}{rgb}{0.21,0.49,0.74}
\usepackage[pagebackref,breaklinks,colorlinks,allcolors=cvprblue]{hyperref}

\def\paperID{19833} 
\def\confName{CVPR}
\def\confYear{2026}

\title{Enhancing End-to-End Autonomous Driving with Risk Semantic Distillaion from VLM}

\author{
Jack Qin\textsuperscript{1}, 
Zhitao Wang\textsuperscript{2}, 
Yinan Zheng\textsuperscript{1}, 
Keyu Chen\textsuperscript{1}, 
Yang Zhou\textsuperscript{2}, 
Yuanxin Zhong\textsuperscript{2}, 
Siyuan Cheng\textsuperscript{2}\thanks{Corresponding author. Email: chengsiyuan8@huawei.com} \\
{\small\textsuperscript{1}Tsinghua University} \quad
{\small\textsuperscript{2}2012 Laboratories, Huawei Technologies} \\
}

\begin{document}
\maketitle
\begin{abstract}
The autonomous driving (AD) system has exhibited remarkable performance in complex driving scenarios. However, generalization is still a key limitation for the current system, which refers to the ability to handle unseen scenarios or unfamiliar sensor configurations.
Related works have explored the use of Vision-Language Models (VLMs) to address few-shot or zero-shot tasks. While promising, these methods introduce a new challenge: the emergence of a hybrid AD system, where two distinct systems are used to plan a trajectory, leading to potential inconsistencies. Alternative research directions have explored Vision-Language-Action (VLA) frameworks that generate control actions from VLM directly. However, these end-to-end solutions demonstrate prohibitive computational demands. To overcome these challenges, we introduce Risk Semantic Distillation (RSD), a novel framework that leverages VLMs to enhance the training of End-to-End (E2E) AD backbones. By providing risk attention for key objects, RSD addresses the issue of generalization. Specifically, we introduce RiskHead, a plug-in module that distills causal risk estimates from Vision-Language Models into Bird’s-Eye-View (BEV) features, yielding interpretable risk-attention maps.
This approach allows BEV features to learn richer and more nuanced risk attention representations, which directly enhance the model’s ability to handle spatial boundaries and risky objects.
By focusing on risk attention, RSD aligns better with human-like driving behavior, which is essential to navigate in complex and dynamic environments. Our experiments on the Bench2Drive benchmark demonstrate the effectiveness of RSD in managing complex and unpredictable driving conditions. Due to the enhanced BEV representations enabled by RSD, we observed a significant improvement in both perception and planning capabilities.
\end{abstract}    

\section{Introduction}
\label{sec:intro}

End-to-end autonomous driving systems have gained significant attentions in both academia~\citep{hu2023planning, jiang2023vad} and industry~\citep{hwang2024emma}. However, their ability to generalize to rare and challenging long-tail scenarios in real-world environments remains limited. To address this, researchers have explored leveraging vision-language models (VLMs) to enhance environmental perception and reasoning. Current approaches primarily adopt two strategies: (1) A dual-system framework~\citep{tian2024drivevlm}, where an upstream VLM provides high-level guidance and a downstream end-to-end model generates trajectories. While this design balances computational efficiency with the strong generalization of VLMs, the decoupled architecture often leads to suboptimal performance due to the "split-brain" problem. (2) Directly fine-tuning VLMs into vision-language-action models~\citep{hwang2024emma} to output trajectories, which requires extensive training data and computational resources, and suffer from slow inference speeds, making them impractical for real-world deployment.


To address these challenges, we propose Risk-Semantic Distillation (RSD), a plug-and-play framework that extracts risk-aware perception capabilities from large VLMs and transfers them to downstream end-to-end driving models. This approach not only enhances generalization in long-tail scenarios but also circumvents the high computational demands of deploying VLMs in real-world systems. Specifically, our framework begins with processing raw image inputs with OV-DINO~\citep{wang2024ov}, a visual grounding model that performs open-vocabulary semantic segmentation. Unlike prior methods that require full fine-tuning of VLMs~\citep{tian2024drivevlm}, our method leverages zero-shot reasoning capability of pretrained VLMs. By designing tailored prompts and a risk-scoring mechanism, our model achieves precise localization of high-risk traffic participants without additional model retraining.

Building upon this framework, we distill the knowledge from large VLMs into an end-to-end driving system to enable efficient real-time deployment. Specifically, we introduce a Risk Prediction Head operating on BEV features~\citep{li2024bevformer}. The key innovation lies in our cross-view feature alignment mechanism: the model back-projects BEV features to perspective view (PV) space first, and then extracts risk-aware semantics via deformable attention~\citep{xia2022vision} and nearest-neighbor matching. This design enables direct 2D image-based supervision while preserving spatial coherence across views, ensuring robust risk prediction without compromising inference speed.

We evaluate our Risk-Semantic Distillation (RSD) framework on the Bench2Drive benchmark, a challenging testing ground for autonomous systems in complex and unpredictable driving scenarios. The results demonstrate that RSD significantly enhances both perception robustness and motion planning safety compared with baseline methods. The main contributions are as follows:
\begin{itemize}
    \item We propose a novel pipeline specifically designed for VLM based risk detection in autonomous driving scenarios, which successfully achieves zero-shot capability without requiring model fine-tuning.
    \item We develop a specialized distillation architecture that effectively transfers risk perception knowledge from large VLMs to compact end-to-end driving models while maintaining computational efficiency.
    \item Our approach demonstrates measurable improvements in both planning stability and perception accuracy compared to conventional methods.
\end{itemize}

\section{Related Work}
\label{sec:formatting}

\paragraph{End-to-End Autonomous Driving}
End-to-end autonomous driving unifies perception, prediction, and planning into a single framework~\citep{chen2024vadv2,hu2023planning,jia2023driveadapter, zheng2025diffusionplanner, jiang2023vad}, the basic idea is to build a fully differentiable learning system that maps the raw sensor input directly into a control signal or a future trajectory. Although these methods have shown promising results, their performance degrades in challenging, long-tail events~\citep{grigorescu2020survey, caesar2021nuplan, nuscenes2019}. This highlights a training gap in current E2E models, which rely soley on the trajectory supervision as sequences of points, lacking the reasoning information necessary for learning rich and robust feature representations to achieve better driving performance. 

\paragraph{Foundation Models for Autonomous Driving}
The application of vision-language models (VLMs) in autonomous driving~\citep{wen2023dilu, renz2024carllava} has evolved along two main directions: dual-system architectures and end-to-end vision-language-action (VLA) models, each with distinct implementations and inherent challenges. The dual-system approach~\citep{tian2024drivevlm, chen2024asynchronous, doll2024dualad} employs a VLM for high-level scene understanding while maintaining a separate planning module for trajectory generation~\citep{sima2024drivelm,zheng2024doe}. This design aims to leverage VLMs' semantic reasoning capabilities for improved perception in complex scenarios~\citep{yang2023llm4drive}. However, it fundamentally suffers from system inconsistency, where the disconnect between the VLM's symbolic reasoning and the planner's geometric processing can lead to conflicting decisions. The asynchronous operation between modules often results in misaligned representations and delayed responses. Alternatively, VLA models~\citep{hwang2024emma, arai2025covla, zhou2025opendrivevla} attempt to unify perception and action prediction by directly mapping visual inputs to control outputs through VLMs~\citep{brohan2023rt,kim2024openvla, team2024octo}. While this end-to-end paradigm eliminates inter-module mismatches, it introduces significant computational inefficiency. The models require extensive adaptation of large VLMs, leading to heavy training demands and impractical deployment requirements. Moreover, their inference speed remains insufficient for real-time driving applications~\citep{fan2018baidu} due to the inherent complexity of VLMs.







\section{Methods}

\begin{figure*}[t]
  \centering
   \includegraphics[width=\linewidth]{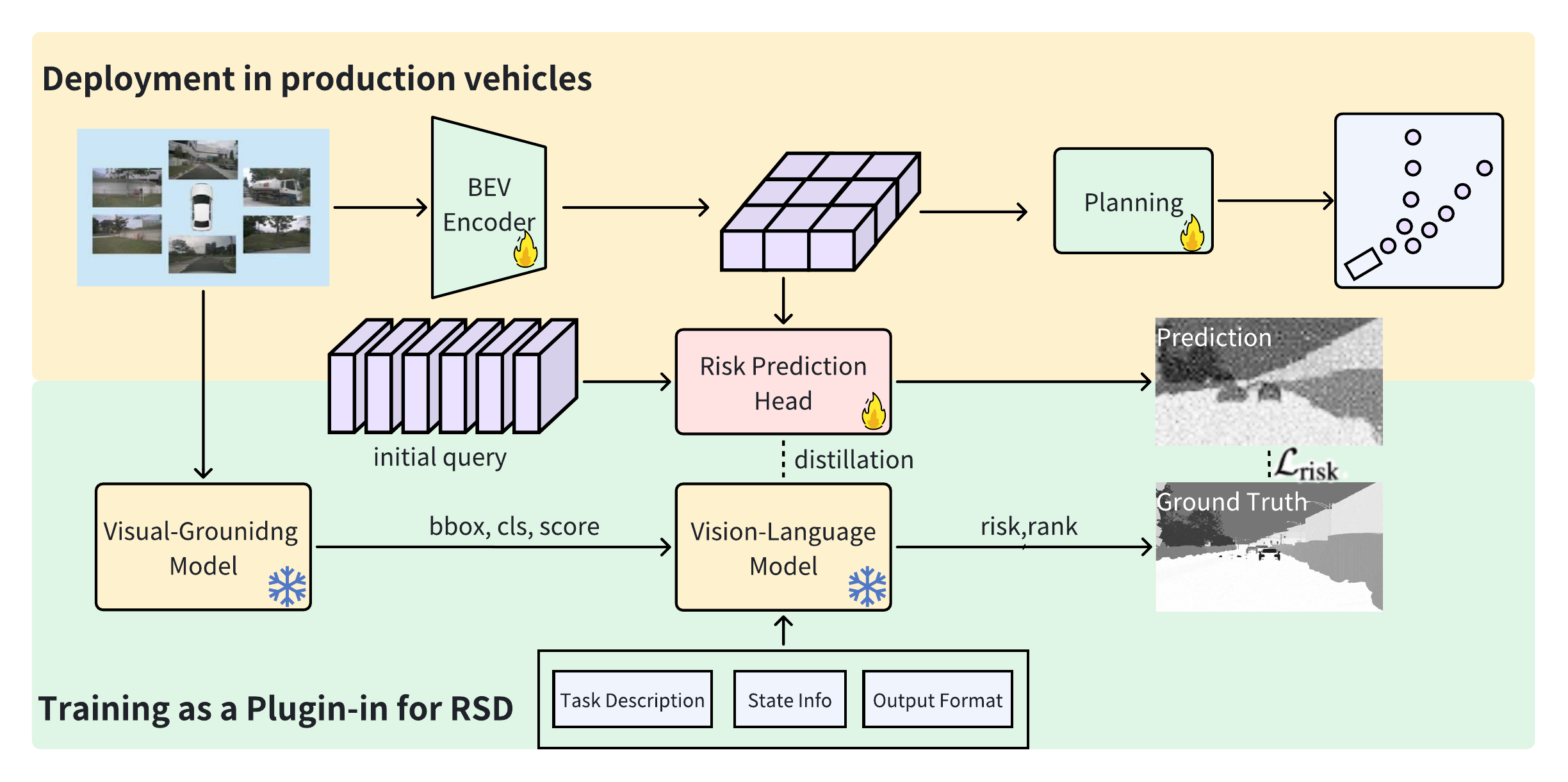}
   \caption{The framework of Risk Semantic Distillation. The diagram illustrates the integration of an End-to-End (E2E) architecture into production vehicles, utilizing the Visual-Language Model (VLM) for risk object identification and subsequent distillation. The VLM processes visual and textual information to identify and categorize critical objects, which are then used to enhance the E2E framework through distillation. This process improves the vehicle's ability to recognize and prioritize risk objects, enabling better risk attention capabilities. As a result, the system significantly enhances the safety of autonomous driving by ensuring the vehicle can detect and respond to potential risks more effectively.}
   \label{fig:DriverAdapter}
\end{figure*}
\subsection{Overview}
As shown in Fig.~\ref{fig:DriverAdapter}, we propose a plug-in, high-generalization, and low-cost end-to-end autonomous driving distillation framework Risk Semantics Distillation(RSD). The RSD efficiently distills common-sense reasoning capabilities regarding the risk semantics of traffic participants from Vision-Language Models (VLMs) into end-to-end autonomous driving networks. Specifically, we leverage the BEV (Bird's Eye View) representations from BEVFormer to predict the ranking and scores of key risky objects from the PV (Perspective View), which directly aligns with the output of the Risk Prediction from VLM. Using the plug-in training structure with strong interpretability and low cost, we effectively improve performance in various long-tail scenarios. Experimental results show significant improvements in both BEV representation qualities and planning performance.

\subsection{VLM enhanced risk semantic Annotation}
\begin{figure}
    \centering
    \includegraphics[width=\linewidth]{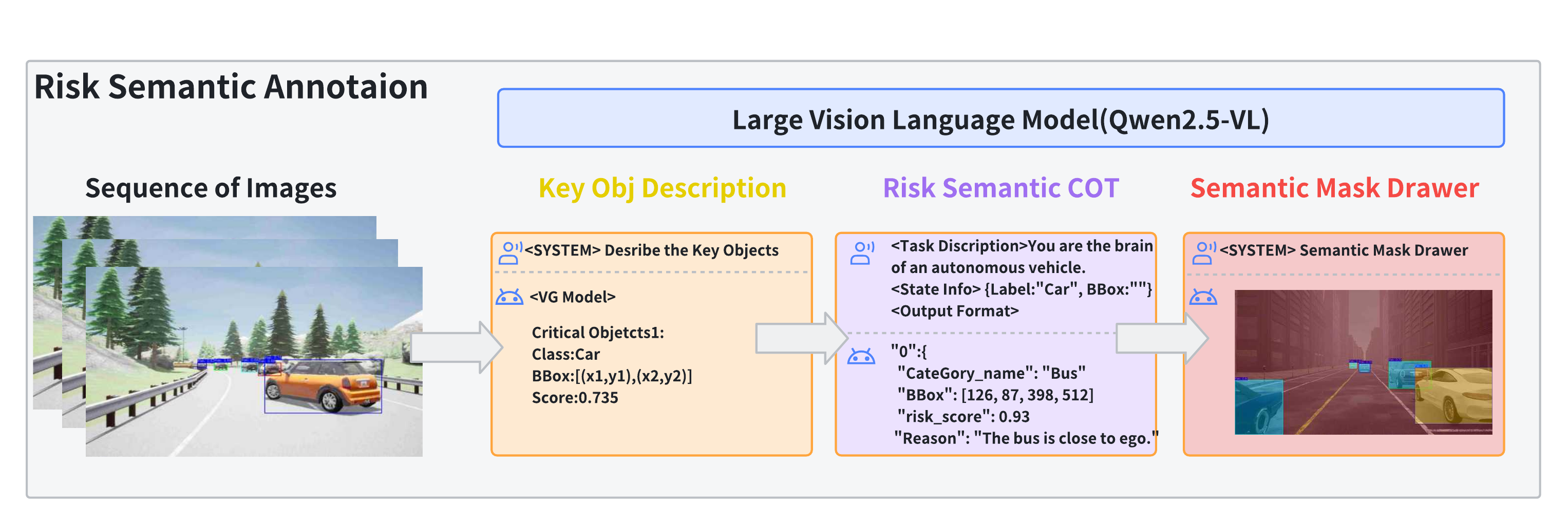}
    \caption{VLM-enhanced Risk Semantic Annotation. A sequence of images is processed through key object description, risk semantic Chain-of-Thought (COT), and a semantic mask drawer. The sequence begins with the extraction of critical objects (such as cars), followed by generating a risk score and reasoning using COT to assess proximity and risk levels. The semantic mask drawer is then used to visualize the detected objects and their associated risk annotations.}
    \label{RSA}
\end{figure}

Figure.~\ref{RSA} presents a detailed workflow of the VLM (Vision-Language Model) enhanced Risk Semantic Annotation process, which leverages a sequence of images processed through various stages to annotate and analyze the potential risk in autonomous driving environments. The input consists of a series of images captured from a sensor system in an autonomous vehicle, containing critical objects that need to be identified and analyzed. We use the Large Vision Language Model (VLM) to describe key objects in the scene. In this stage, the VLM model identifies and provides descriptions of critical objects, such as vehicles, pedestrians, or obstacles in the images. The Visual-Grounding (VG) Model is employed to describe the objects with specific details such as their classes (e.g., "Car") and bounding box coordinates, which mark the location of the object in the image. The model outputs a score that indicates the confidence level of the detection (e.g., a score of 0.735). After that, Risk Semantic CoT (Chain of Thought) is used to interpret the risk associated with the identified objects. The system generates task descriptions that focus on the state of the objects related to the autonomous vehicle. For example, it evaluates if a car is moving too close to the vehicle or if an object is in a hazardous position. The system outputs the risk score and provides reasoning, such as "The bus is close to ego," where the ego refers to the autonomous vehicle. This analysis is critical in assessing potential risks posed by the environment. Finally, Semantic Mask Drawer is used, which visually highlights the identified risk in the scene. Using the bounding boxes and descriptions from the previous steps, this tool draws semantic masks over the relevant objects in the image. These masks help the system visually demonstrate areas of risk, aiding in the process of risk analysis and decision-making.

\subsubsection{Risk Semantic Annotation}
\label{Risk Semantic Annotation}

\paragraph{Visual Grouding State Info Genaration}
We utilize the OV-DINO\cite{wang2024ov} model, an open-vocabulary object detection framework that integrates both visual and textual information to identify arbitrary objects based on natural language descriptions, to obtain category labels and bounding boxes from frame-by-frame images in the original data. The results are in a fixed JSON format, serving as crucial input for the VLM model. This approach compensates for the VLM model's limited spatial grounding capabilities without requiring pre-training focused on ranking and scoring hazardous objects. We adopt an open vocabulary approach that includes commonly observed traffic participants and scene elements. Specifically, the vocabulary is defined as:
\{\text{pedestrian}, \text{car}, \text{bus}, \text{bicycle}, \text{motorcycle}, \text{truck}, \text{fence}, \text{barrier},  \text{construction cone}, etc.\} This vocabulary remains flexible and can be extended to accommodate additional domain-specific objects as needed.

\paragraph{Prompt Learning for Risk Semantic Annotation using VLM}
Based on the VG model, we use the output of the VG model as context input, and simultaneously provide the forward view input. The Qwen model is then used to output the ranking and scores of key risk objects in the image. We propose a risk consistency metric, Diff\_Risk, to evaluate the labeling performance of the Risk Semantic Annotation framework. The metric is defined as the mean Frobenius distance between the ground truth (GT) risk prediction and the VLM-predicted risk prediction. Mathematically, it can be expressed as follow:

\begin{equation}
    \text{Diff\_Risk} = \frac{|R_{\text{GT}} - R_{\text{VLM}}|}{\|R_{\text{GT}}\|_F}
\end{equation}

where
\( R_{\text{GT}} \) represents the ground truth risk scores,
\( R_{\text{VLM}} \) represents the predicted risk scores by VLM,
\( \|R_{\text{GT}}\|_F \) is the Frobenius norm of the ground truth risk scores. The Frobenius norm \( \|R_{\text{GT}}\|_F \) is used for normalization, ensuring that the metric is dimensionless and can be applied consistently across different datasets and risk scales.

In this study, we evaluate four methods based on their ability to annotate and score critical risk objects. The first method, VLM, directly utilizes the Qwen-2.5 model to perform object recognition and scoring for key risk objects. The second method, Enhanced VLM (Prompt), improves the model's performance by adjusting the prompt, specifically optimizing the Task Description and Output Format, which results in higher scoring accuracy. The third method, Enhanced VLM (Prompt + VG), further enhances the system by incorporating a Visual Grounding (VG) model into the State Info module, leading to improved performance by providing more precise object localization and context understanding. Finally, Enhanced VLM (Prompt + VG + Risk-Level COT) introduces a Risk-Level Chain-of-Thought (CoT) during the reasoning process, which incorporates a qualitative assessment of object risk, thereby strengthening the VLM’s ability to focus on risk-related features and improving the overall accuracy of the risk annotations.

\begin{table}
  \caption{Risk Semantic Annotation}
  \label{Tab: Risk Semantic Annotation}
  \centering
  \begin{tabular}{@{}lc@{}}
    \toprule
    Method & Diff\_Risk$\downarrow$\\
    \midrule
    Base VLM(Qwen-2.5) &$0.45\pm0.05$ \\
    Enhanced VLM(Prompt) & $0.36\pm0.05$ \\
    Enhanced VLM(Prompt + VG) & $0.23\pm0.03$ \\
    Enhanced VLM(Prompt + VG + Risk)& \textbf{$0.15\pm0.02$}\\
    \bottomrule
  \end{tabular}
\end{table}



As shown in Tab.~\ref{Tab: Risk Semantic Annotation}, the result presents a comparison of different methods for Risk Semantic Annotation based on the Mean Frobenius Distance, where lower values indicate better performance. The baseline VLM (Base) method achieves a distance of 0.45. By enhancing the model with Prompt-Enhanced VLM, the performance improves, reducing the distance to 0.36. Further enhancement through VG-Enhanced VLM leads to a more significant improvement, with a distance of 0.23. The best performance is achieved with the Risk-Level CoT, which produces the lowest Mean Frobenius Distance of 0.15, demonstrating the highest accuracy in risk annotation. This progression highlights the effectiveness of incorporating additional enhancements to improve the model's precision in risk assessment tasks.

\subsection{Risk Semantic Distillation}
The core of the RSD framework lies in enhancing BEV representations by introducing a new risk object prediction task.  This task focuses on predicting the risk score of traffic objects that exhibit high uncertainty or potential danger from a bird's-eye view.  The Risk Prediction Head generates both a ranking and a risk score for each object. These predictions are then aligned with the outputs of a Visual-Language Model (VLM), which addresses risk semantics related to traffic objects. In our approach, we develop an auxiliary risk semantic prediction head, which takes scenario-specific features as input.  This enables the model to extract valid causal inferences from the VLM’s responses, especially in long-tail scenarios, such as semi-occlusions and blind zones, that are often challenging for conventional models.  As a result, the enhanced perception module feeds more relevant information into the planning module, allowing it to better focus on and address risk objects, ultimately improving the overall planning outcomes.

\begin{figure*}[!t]
  \centering
\includegraphics[width=\linewidth]{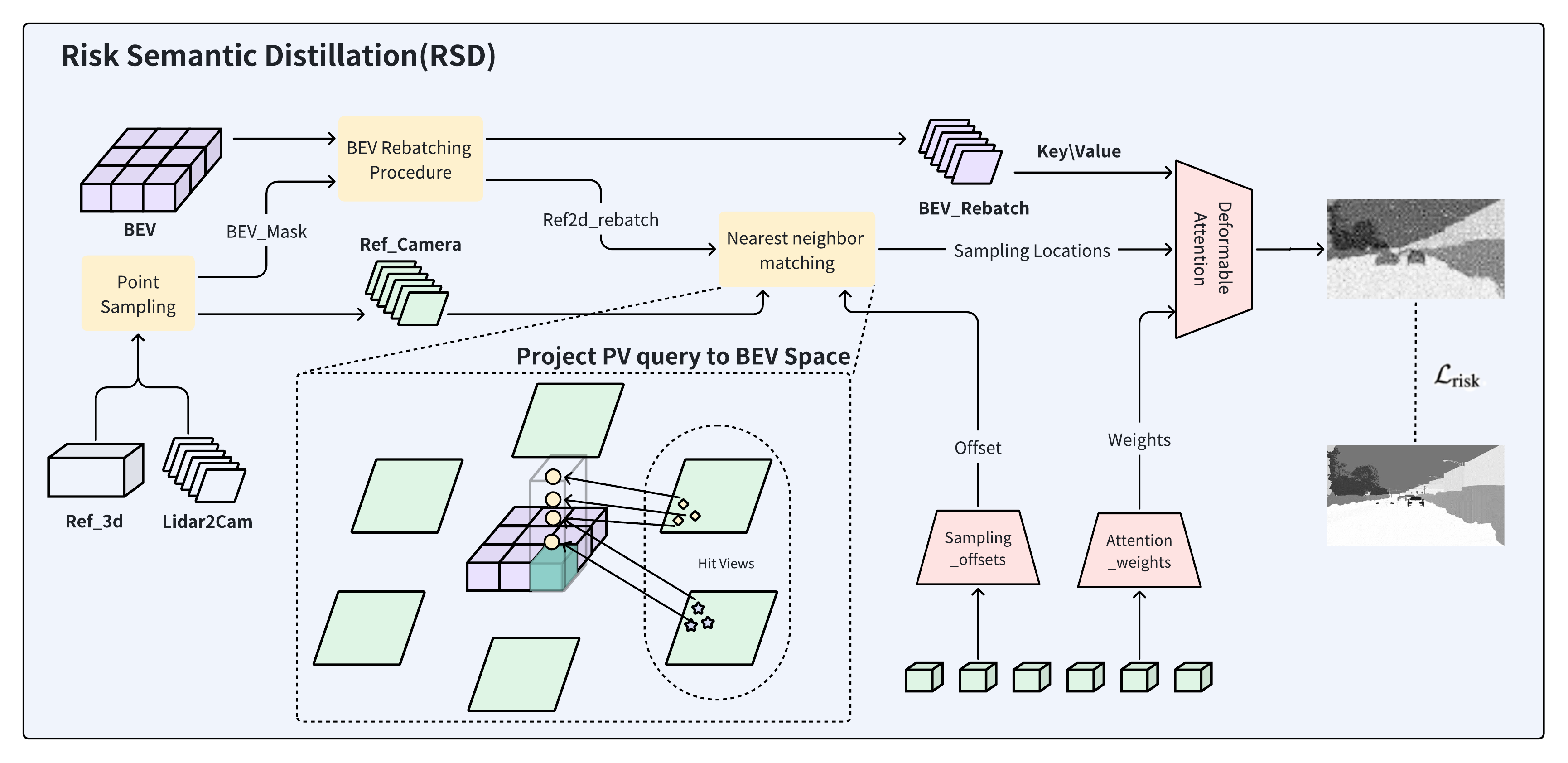}
   \caption{The framework of Risk Semantic Distillation (RSD). This figure illustrates the process of projecting the Perspective View (PV) query into the BEV space, followed by the BEV re-batching procedure and nearest neighbor matching. It showcases the integration of point sampling, BEV masking, and the alignment of the reference camera with the 3D lidar-to-camera matrix. The system utilizes deformable attention for optimizing the risk semantic information extraction, with sampling offsets, attention weights, and locations playing crucial roles in enhancing the precision of risk object prediction. The final output is used for accurate decision-making in autonomous driving systems.}
   \label{fig:RiskHead}
\end{figure*}

\paragraph{BEV Rebatching Procedure}
In multi-view perception systems, it is computationally inefficient to allow all camera views to attend to all Bird's Eye View (BEV) queries\cite{vaswani2017attention}. For a BEV feature, we must have a special BEV mask and rebatch process, that is, the projected 2D point can only fall on some of the six PV views, while the other views are not hit. Here, we call the hit view $V_{hit}$\cite{li2024bevformer}. To improve memory efficiency and inference speed, we introduce a rebatching procedure\cite{li2024bevformer} in which each camera only processes those BEV queries that are within its field of view. 

Let $\mathcal{BEV} \in \mathbb{R}^{B \times N_{BEV} \times d}$ denote the original BEV query tensor, and let $\mathcal{R}_{2d} \in \mathbb{R}^{ B  \times N_{\text{cam}} \times N_{BEV} \times D \times 2}$ be the projected 2D reference points. We first extract the set of valid BEV queries for each camera:
\begin{equation}
    \mathcal{I}_{k}^{(b)} = \left\{ q \mid \sum_{d=1}^{D} \text{BEV\_mask}_{k}^{(b)}[q, d] > 0 \right\}
\end{equation}
Where q denotes the position of BEV\_mask.
Let $L_{\max} = \max_{k, b} |\mathcal{I}_{k}^{(b)}|$ be the maximum number of visible queries across all views and batches. We initialize rebatching tensors:
\(
\mathcal{BEV}' \in \mathbb{R}^{B \times N_{\text{cam}} \times L_{\max} \times d}, \quad
\mathcal{R}'_{2d} \in \mathbb{R}^{B \times N_{\text{cam}} \times L_{\max} \times D \times 2}
\). For each batch $b$ and camera $k$, we insert the visible queries $q_{vis}$:
\begin{equation}
    \begin{aligned}
        & \mathcal{BEV}'[b, k, l] = \mathcal{BEV}[b, q_{vis}], \\
        & \mathcal{R}'_{2d}[b, k, l, d] = \mathcal{R}_{2d}[d, b, q_{vis}, k]\\
        &\forall q_{vis} \in \mathcal{I}_{k}^{(b)}, \quad l = \text{index}(q_{vis})
    \end{aligned}
\end{equation}

\paragraph{Nearest Neighbor Matching}
We first lift each query on the BEV plane to a pillar-like query, and sample \( N_{ref3D} \) reference points from the pillar\cite{lang2019pointpillars}. These points are projected into 2D views,  \(
R_{cam} = \text{lidar2img}(R_{ref3d})
\), where \( R_{\text{ref3d}} \) represents the reference points from the 3D data. This approach ensures that for each batch sample \( b \), camera \( k \), and query \( q \), you select the most appropriate 2D reference point (from the projected BEV queries $\mathcal{BEV}'$) using the nearest neighbor in the image plane.  The resolution of the pixels of PV queries is set to \( 6 \times 80 \times 45 \), which exceeds the maximum value observed in the hit views \( V_{\text{hit}} \) after the rebatching of \( \text{ref}_{2d} \). To ensure proper matching, nearest-neighbor matching is performed between \( \text{ref}_{2d} \) and \( \text{ref}_{\text{camera}} \) to find the closest points in the \( \text{ref}_{2d} \) matrix:
\begin{equation}
\begin{aligned}
    &D_{b, k, q, l} = \sqrt{(x_{\text{cam}}^{b, k, q} - x_{2d}^{b, k, l})^2 + (y_{\text{cam}}^{b, k, q} - y_{2d}^{b, k, l})^2}\\
    &l_{\text{NN}} = \arg\min_l \left( D_{b, k, q, l} \right) \quad \mathcal{R}'_{2d}[b, k, l_{\text{NN}}, :] = \mathcal{R}_{2d}[b, k, q, :]
\end{aligned}
\end{equation}

Next, the matched 2D points are considered as the reference points for the BEV features, and features are sampled from the hit views \( V_{\text{hit}} \) around these reference points.

\paragraph{Deformable Attention}
As shown in Fig.~\ref{fig:RiskHead}, we get BEV feature as keys and values $\mathcal{BEV}'$, the initialized six learnable risk PV features as queries $Q_{pv}$. The BEV features serve as the key and value space, where the sampled values are weighted using an deformable attention mechanism\cite{zhu2020deformable} to produce the final result.

\begin{equation}
\begin{aligned}
    & \text{RHA}(Q_{pv}, \mathcal{BEV}') = \\
    & \frac{1}{|V_{\text{hit}}|} \sum_{i \in V_{\text{hit}}} \sum_{j=1}^{N_{\text{ref}}} \text{DeformAttn}(Q_{pv}, P(p, i, j), (\mathcal{BEV}')^i)
\end{aligned}
\end{equation}


where $i$ indexes the camera view, $j$ indexes the reference points, and $N_{ref}$ is the total reference points for each risk semantic camera query. $(\mathcal{BEV}')^i$ is the features of $(\mathcal{BEV}$ matching for the i-th camera view. For each risk semantic camera query $Q_{pv}$, we use a
project function $P(p, i, j)$ to get the j-th reference point of $BEV$ on the i-th camera view image.

\subsection{Vision-based End-to-end Model}
The model architecture is based on the VAD\cite{jiang2023vad}, which integrates visual, motion, and map-based features. It incorporates a ResNet-50 backbone for extracting image features, and uses FPN (Feature Pyramid Network) for multi-scale feature aggregation. The key components for trajectory prediction are transformer-based decoders, such as CustomTransformerDecoder, which handle motion, map, and object detection tasks. The output of perception encoder, BEV feature, aggregates all information from upstream modules and directs downstream trajectory heads to output valid trajectory information with causal inference properties. The framework of RiskHead, shown in Fig.~\ref{fig:RiskHead}, is designed to process Bird's Eye View (BEV) data, extract relevant features, and make risk predictions to guide planning decisions, the loss is as followed:
\begin{equation}
\begin{aligned}
    \mathcal{L} = &\omega_1 \mathcal{L}_{\text{map}} + \omega_2 \mathcal{L}_{\text{mot}} + \omega_3 \mathcal{L}_{\text{col}} + 
\omega_4 \mathcal{L}_{\text{bd}} + \\
     &\omega_5 \mathcal{L}_{\text{dir}} + \omega_6 \mathcal{L}_{\text{limi}} + 
\omega_7 \mathcal{L}_{\text{risk}}.\\
\mathcal{L}_{\text{risk}} = &\left\| R_{\text{pred}} - R_{\text{gt}} \right\|_1,
\end{aligned}
\end{equation}
The overall loss function \( \mathcal{L} \) is composed of multiple weighted components, each targeting a specific aspect of autonomous driving behavior, such as map adherence, motion smoothness, collision avoidance, and boundary constraints. To enhance risk-awareness, we introduce an additional term, \( \mathcal{L}_{\text{risk}} \), which captures the discrepancy between predicted risk semantic $R_{\text{pred}}$ and ground-truth risk semantic $R_{\text{gt}}$ using the L1 norm. This term encourages the model to accurately estimate the relative risk levels of traffic participants, thereby improving safety-critical decision-making, especially in complex or long-tail scenarios.

\section{Experimental Settings}
\label{Experimental Settings}
\subsection{Dataset}
we evaluate the method on the Bench2Drive dataset\cite{jia2024bench2drive}, a large-scale antonomous driving benchmark consists of 44 long-tail corner scenarios, including high-risk situations such as Cutin, Construction, Accident, HardBrake, Merge, and other critical events commonly encountered in autonomous driving environments. This dataset is specifically designed to address the challenges posed by the long-tail distribution of data, both from perception and behavior perspectives, which is a significant bottleneck in the development and validation of autonomous driving systems. While most common datasets like nuScenes predominantly feature the ego vehicle driving in straight paths, representing only a narrow subset of potential real-world scenarios. The bench2drive has a large-scale expert dataset with comprehensive
annotations including 3D bounding boxes, depth, and semantic segmentation, sampled at 10 Hz, to
serve as the official training set. The dataset features a 360° multicamera rig composed of six synchronized cameras (front,
front-left, front-right, back, back-left, back-right) with minimal
field-of-view overlap. Precise camera intrinsic and extrinsic are
provided for each frame to ensure accurate spatial alignment.
\subsection{Metrics}
We used common perception and planning metrics to validate the effectiveness and rationality of BEV (Bird's Eye View) representation learning. For perception, we evaluated performance using mean Average Precision (mAP), mean Absolute Trajectory Error (mATE), mean Absolute Spatial Error (mASE), mean Absolute Orientation Error (mAOE), mean Absolute Velocity Error (mAVE) and Normalized Discounted Score (NDS). For planning, we used Average Displacement Error (ADE) and plan\_obj\_box\_col, a collision indicator. These metrics provide a comprehensive evaluation of the BEV representation's ability to detect objects and plan safe, accurate trajectories in autonomous systems. 

\subsection{Implementation Details}
\label{Implementation Details}
The model is specifically designed for trajectory prediction and object detection within the context of autonomous driving. The configuration outlines several crucial parameters, including the utilization of past and future frames for trajectory forecasting, as well as the resolution of the bird-eye-view (BEV) feature map. Notably, the model incorporates 2 past frames and 6 future frames, thereby ensuring a comprehensive representation of both historical and predicted vehicle trajectories, thus providing a robust context for modeling both current and future states. The BEV feature map is configured with a resolution of $100\times100$, which facilitates the effective processing and interpretation of spatial information, essential for accurate scene understanding and object detection. Furthermore, the model employs the AdamW optimizer with a learning rate of 2e-4, complemented by a cosine annealing learning rate schedule. This combination ensures a stable and efficient training process, enabling the model to adapt to complex environmental contexts and optimize its performance in trajectory prediction and object detection tasks.

\begin{table*}
  \caption{Perception Results.}
  \label{tab:perception}
  \centering
  \begin{tabular}{ccccccc}
    \toprule
    Method & mAP$\uparrow$ & mATE$\downarrow$ & mASE$\downarrow$  & mAOE$\downarrow$ & mAVE$\downarrow$ & NDS$\uparrow$\\
    \midrule
    VAD &0.5066 & 0.3852 &0.0854 &0.0308 &0.5936 &0.6042\\
    \textbf{VAD-RSD} & \textbf{0.5195} & \textbf{0.3213} & \textbf{0.0544} &\textbf{0.0272} & \textbf{0.5424} &\textbf{0.6280}\\
    \bottomrule
  \end{tabular}
\end{table*}

\begin{table*}
\small
\centering
\caption{Planning RESULTS.}
\begin{tabular}{lcccccccc}
\toprule
{Method} & \multicolumn{3}{c}{plan\_L2(ADE) $\downarrow$} & \multicolumn{3}{c}{plan\_obj\_box\_col(Col) $\downarrow$} \\ 
\cmidrule(lr){2-4} \cmidrule(lr){5-7}
                    & ADE\_1s 
                    & ADE\_2s           
                    & ADE\_3s           
                    & Col\_1s
                    & Col\_2s
                    & Col\_3s           \\ 
\midrule
VAD                
&$0.4541$ 
& $0.9125$  
& $\mathbf{1.4778}$ 
& $0.001017$ 
& $0.002012$ 
& $0.002961$ \\ 
\textbf{VAD-RSD}            
& $\mathbf{0.4023}$ 
& $\mathbf{0.8978}$   
& ${1.5344}$ 
& $\mathbf{0.000509}$ 
& $\mathbf{0.001249}$   
& $\mathbf{0.002267}$ \\ 
\bottomrule
\end{tabular}
\label{tab:Planning}
\end{table*}


\section{Results}
Our experimental evaluation is designed to address the following three key questions: a)  Does RSD lead to improved BEV representation learning? b) Does RSD help the planning module better attend to risk objects and improve planning performance, especially in long-tail scenarios? c) Can the Risk Prediction Head in RSD reconstruct risk semantics and exhibit strong interpretability?

\subsection{Quantitative Results}
\paragraph{Perception}
As shown in Table~\ref{tab:perception}, VAD-RSD shows significant improvements across all six metrics, demonstrating its effectiveness in enhancing perceptual abilities. Especially for  mASE (mean Absolute Spatial Error) metric, which reflects the spatial prediction accuracy, where smaller values indicate better performance. The VAD-RSD model shows a significant improvement with the value decreasing from 0.0854 to 0.0544, indicating a better spatial perception ability.
mAVE (mean Absolute Velocity Error) metric measures the accuracy of velocity prediction. A lower value signifies more accurate velocity estimation. The VAD-RSD model shows a 10\% improvement in velocity prediction, suggesting that the improved ability of the model to handle risk objects has led to a better overall perception performance.

The substantial improvement in the VAD-RSD model can be attributed to the Risk Semantic Distillation (RSD) technique, which continuously focuses on risk objects (objects with higher uncertainty or likelihood of causing a problem). This focus enhances the model's ability to refine its predictions, particularly for challenging tasks such as velocity and spatial estimation. The RSD enables better learning of the Bird's Eye View (BEV) representation, which in turn improves the model's accuracy in velocity and spatial predictions. 

\paragraph{Planning}
As shown in Table ~\ref{tab:Planning}, the ADE results indicate better trajectory prediction accuracy, particularly for shorter time steps. The collision results demonstrate a significant reduction in the predicted collisions for VAD-RSD, particularly over longer time horizons.
   
Substantial improvements in both planning and collision metrics with VAD-RSD can be attributed to the effectiveness of Risk Semantic Distillation (RSD). RSD helps the model focus on high-risk objects and potential hazards, enhancing the model's spatial perception and risk prediction capabilities. As a result, the model's planning accuracy (measured by ADE) and its ability to avoid collisions (measured by Col) are both significantly improved. This improvement mirrors the enhanced perception results seen earlier, where VAD-RSD outperformed VAD in terms of spatial and velocity estimation accuracy. The reduction in collision predictions and improved trajectory planning reflects the model’s strengthened ability to anticipate and avoid risky scenarios, suggesting that the model has learned a more accurate and robust Bird's Eye View (BEV) representation. This, in turn, allows the model to make better informed decisions during planning and collision avoidance, leading to safer and more accurate path predictions.

\paragraph{Closed-loop Metric}

In our closed-loop simulation experiment within the CARLA environment, we trained our model using only a \textbf{tiny subset} of the Bench2Drive dataset—specifically, \textbf{10\%} of the base dataset and \textbf{1\%} of the full dataset. For validation, we employed the \textbf{Dev10 benchmark}\cite{jia2024bench2drive,jia2025drivetransformer}, comprising 10 carefully selected clips from the official 220 routes. These clips were chosen to be both challenging and representative, exhibiting low variance, which is officially recommended for ablation studies to prevent overfitting on the entire Bench2Drive220 routes.

As shown in Tab.~\ref{tab:Closed-Loop Experiment}, the results demonstrate that integrating the RSD plugin into End-to-End backbone significantly enhances the performance of the VAD model.

\begin{table}[ht]
\centering
\caption{Closed-Loop Experiment.}
\begin{tabular}{ccccccc}
\toprule
Method & Driving Score$\uparrow$ & Success Rate$\uparrow$\\
\midrule
VAD-Tiny &36.306 & 0.167\\
\textbf{VAD-Tiny-RSD} & \textbf{46.662} & \textbf{0.278} \\
\bottomrule
\end{tabular}
\label{tab:Closed-Loop Experiment}
\end{table}

\begin{figure*}[htbp]
    \centering
    \begin{minipage}{0.35\textwidth}
        \centering
        \includegraphics[height=2.7cm]{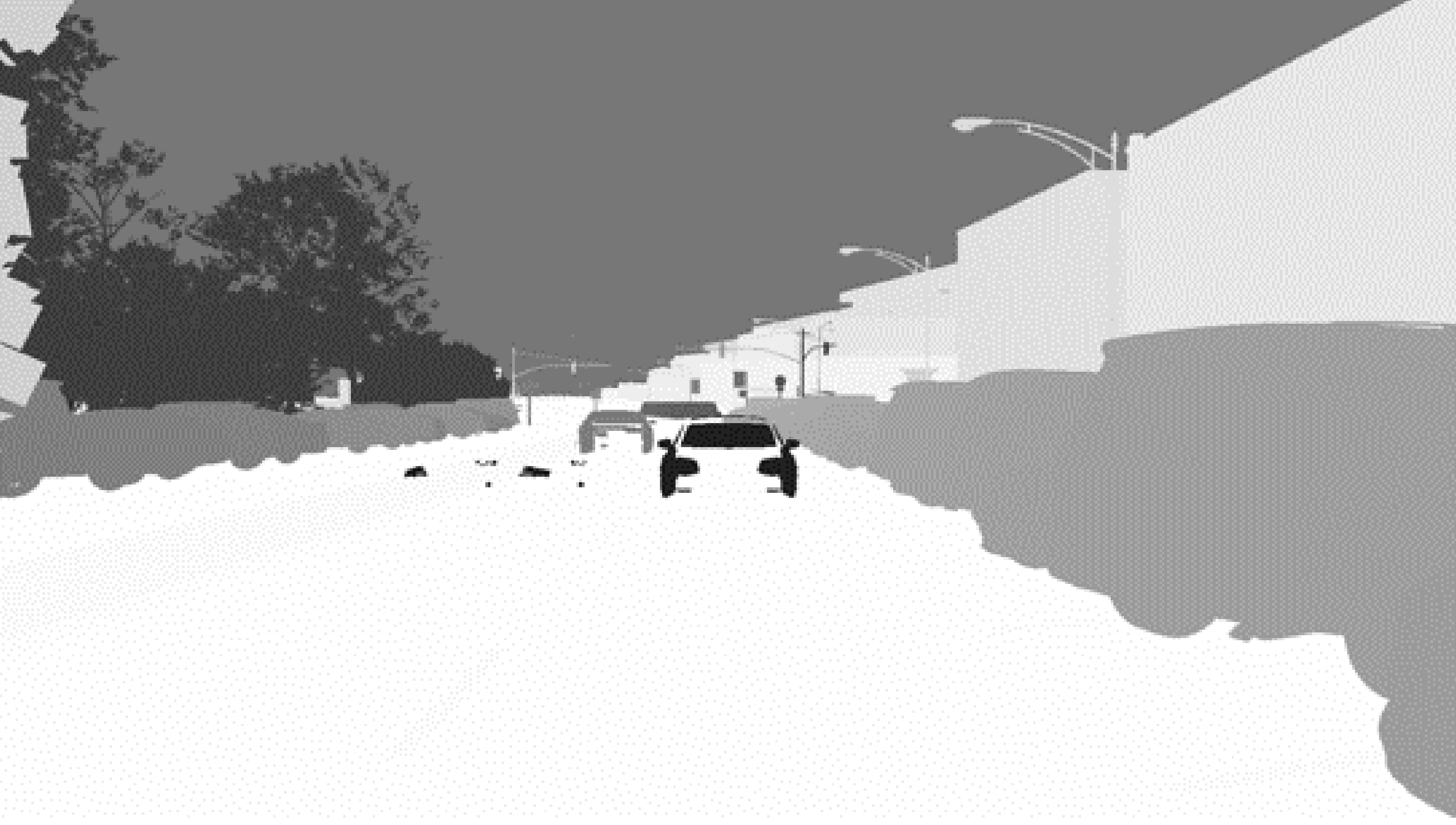}
    \end{minipage}
    \begin{minipage}{0.35\textwidth}
        \centering
        \includegraphics[height=2.7cm]{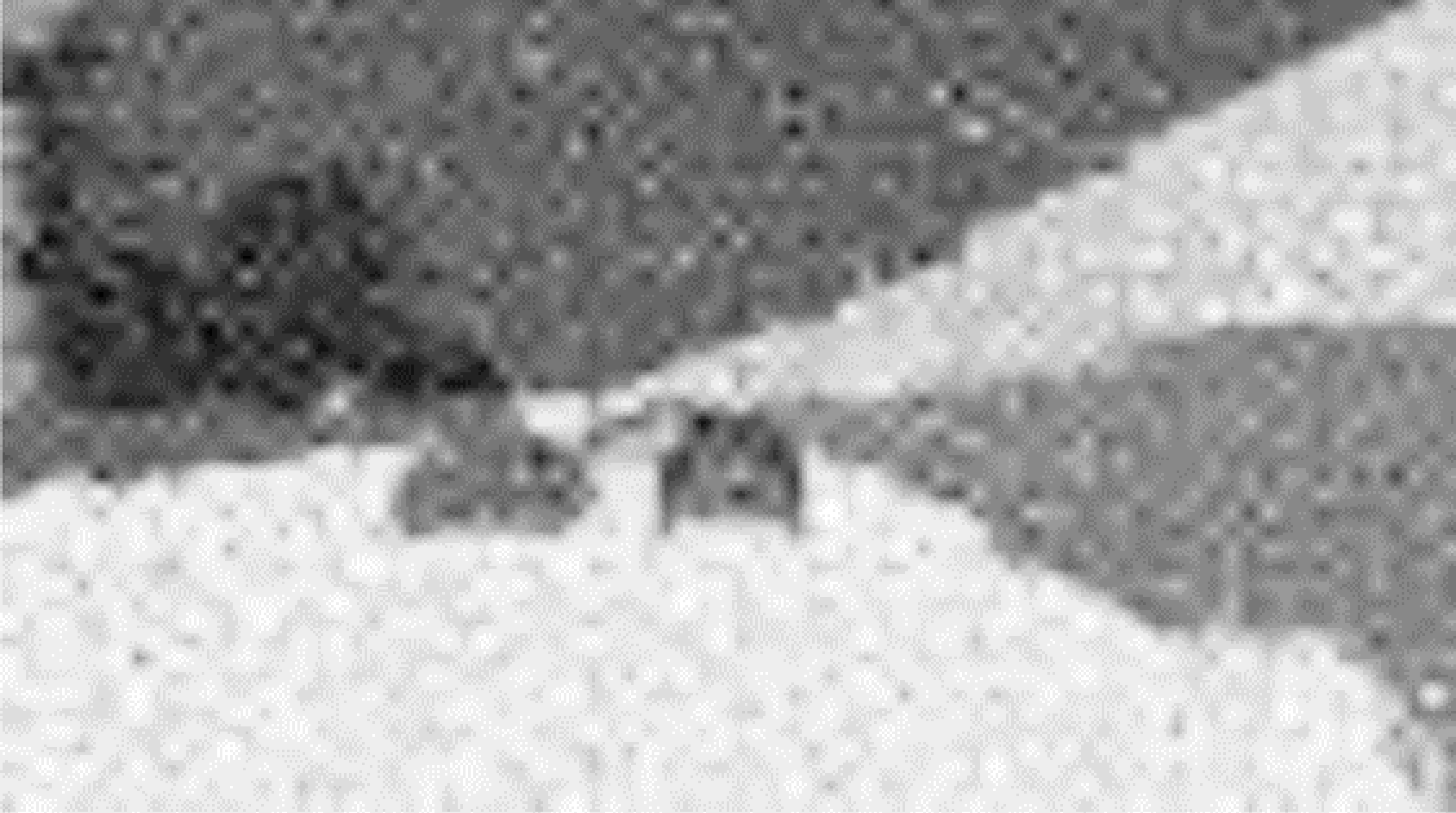}
    \end{minipage}
    \begin{minipage}{0.35\textwidth}
        \centering
        \includegraphics[height=2.7cm]{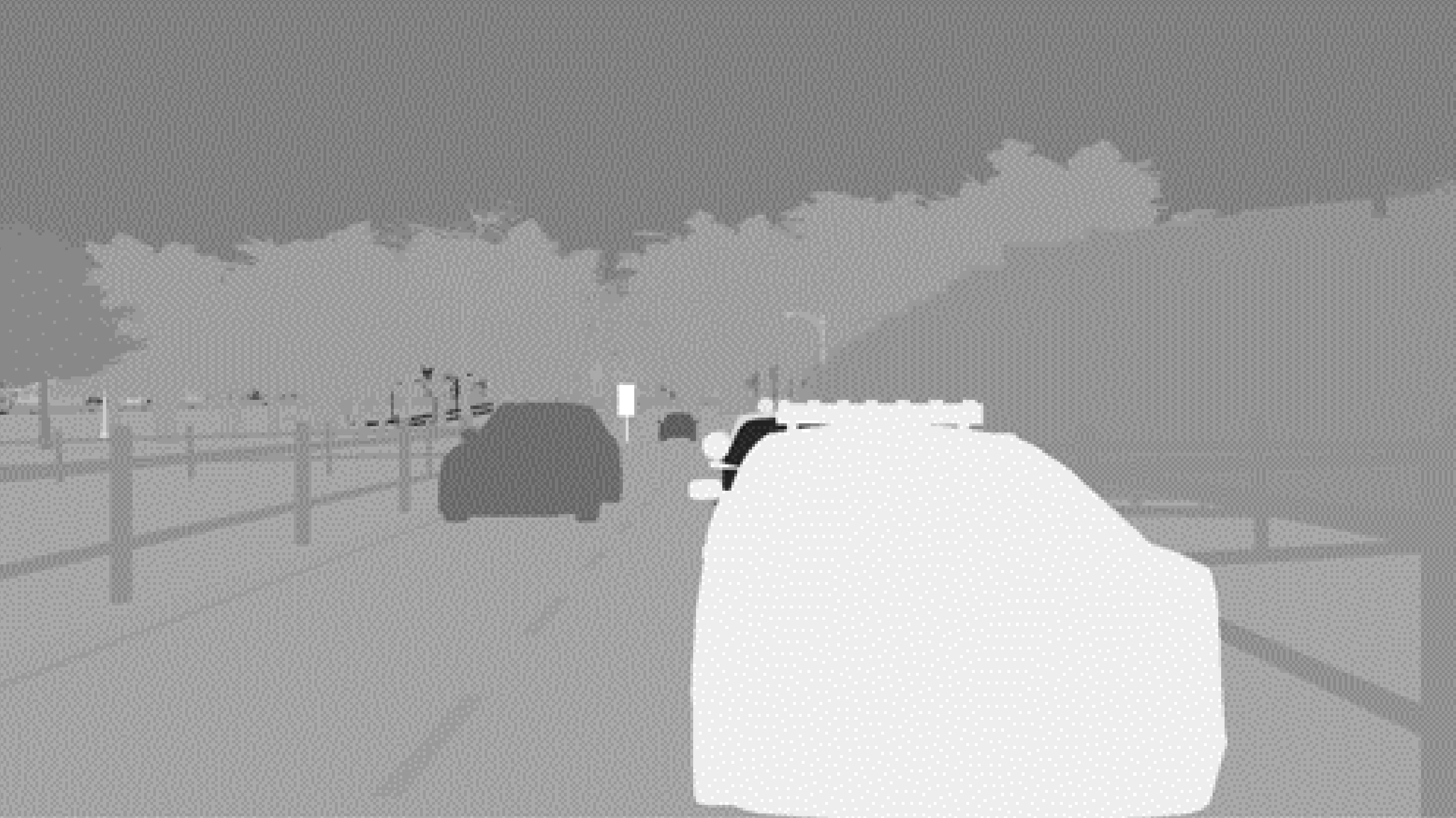}
    \end{minipage}
    \begin{minipage}{0.35\textwidth}
        \centering
        \includegraphics[height=2.7cm]{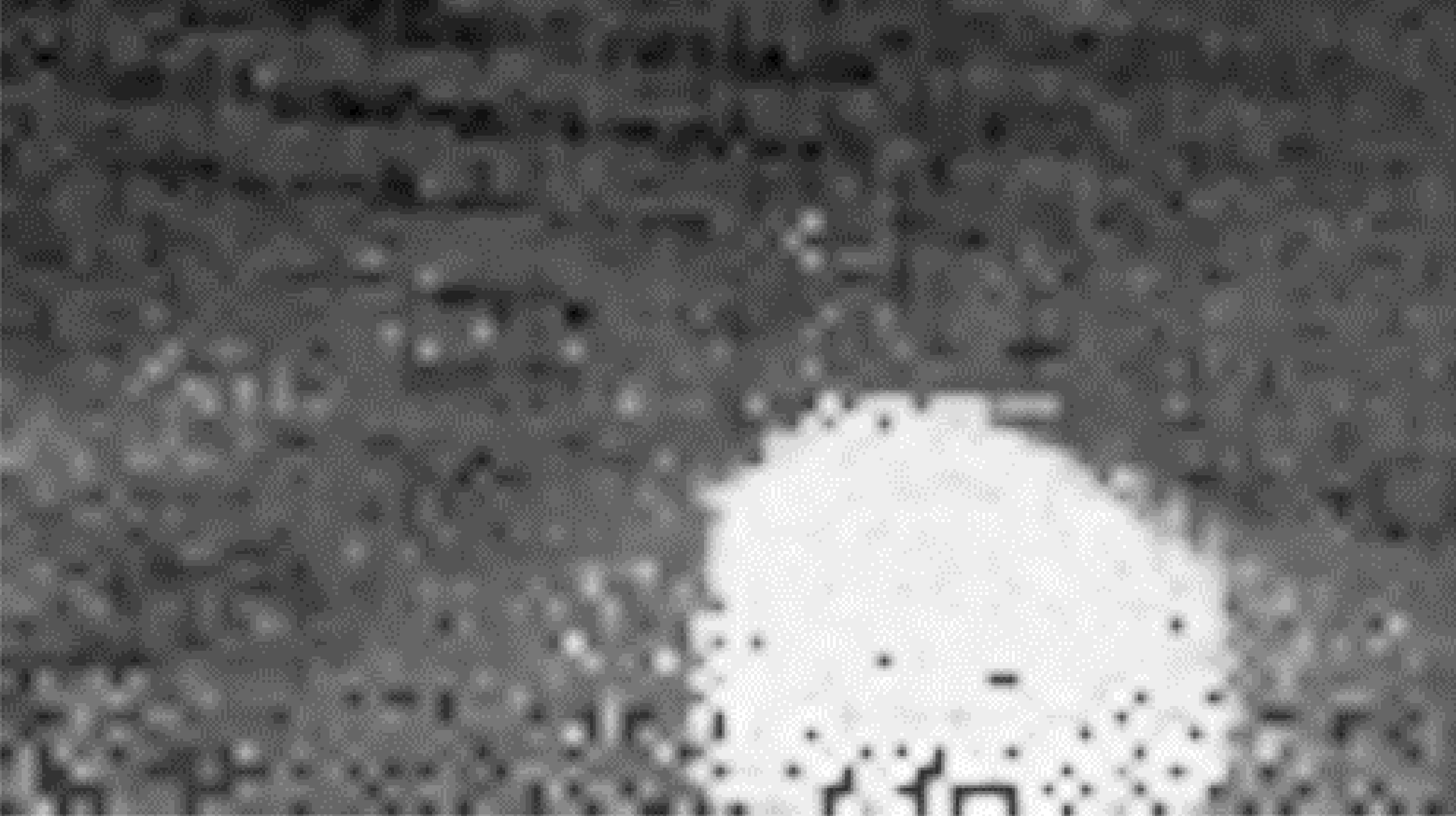}
    \end{minipage}
    \caption{\textbf{Risk Head Reconstruction}. The reconstructed images show how different levels of risk are highlighted from the BEV features, guiding the learning of more effective BEV representations. This process helps the system focus on critical objects and scenarios, improving the vehicle's ability to navigate and make safe decisions in real-world environments.}
    \label{fig:risk Semantic}
\end{figure*}

\subsection{Qualitative Analysis}
As shown in Fig.~\ref{fig:risk Semantic}, the model's ability to highlight critical objects that pose potential risks is evident from the distinct visual patterns observed in the images. The image on the left in each row shows the full scene with general environmental features, while the corresponding image on the right zooms in on the specific objects that are considered key risks. These risk-critical objects are highlighted with higher contrast, indicating their significance in the context of risk prediction. The reconstructions exhibit a high level of resolution, focusing on fine-grained details of objects such as the contours of vehicles or pedestrians. This focus ensures that even subtle risk factors are captured, enabling the model to prioritize these objects during decision-making processes.

By leveraging simple bounding boxes instead of dense pixel-level annotations, our approach significantly reduces the annotation cost and training complexity, while still enabling the model to focus on semantically meaningful, high-risk regions. The visualized results show that the risk head consistently attends to critical objects (e.g., crossing vehicles, occluded cars, or oncoming traffic), validating its ability to extract interpretable and task-relevant risk semantics. 
\section{Conclusion}
This paper proposes a plug-and-play training framework for autonomous driving that effectively leverages the reasoning capabilities of large multimodal models (VLMs) in risk-aware intent recognition. By introducing a novel risk semantic distillation mechanism, the framework transfers the VLM’s understanding of critical and high-risk objects into an end-to-end driving backbone without requiring architectural modifications or additional inference overhead. The approach improves BEV (Bird’s-Eye View) representation learning and leads to significant performance improvements, particularly in long-tail scenarios. Notably, the method achieves this without fine-tuning the VLM or relying on extra annotated datasets, ensuring both efficiency and scalability in practical deployment. 
The model is lightweight, with a total parameter size of approximately 50M, the weight size of which is approximately 1\% of that of typical VLM-AD architectures, resulting in over a 10× improvement in inference efficiency. This lightweight design enables real-time deployment while retaining the essential risk-aware capabilities distilled from large-scale vision-language models. This low-latency and compact design makes VAD-RSD well-suited for real-time autonomous driving applications.

{
    \small
    \bibliographystyle{ieeenat_fullname}
    \bibliography{main}
}


\end{document}